\pdfoutput=1

\documentclass[11pt]{article}

\usepackage{EMNLP2022}

\usepackage{times}
\usepackage{latexsym}

\usepackage[T1]{fontenc}

\usepackage[utf8]{inputenc}
\usepackage{soul}
\usepackage{microtype}

\usepackage{inconsolata}

\usepackage{multirow}

\usepackage[color=green!50]{todonotes}
\usepackage{xspace}
\newcommand{\meddra}{MedDRA\xspace}

\newcommand{\pubmedbert}{PMB\xspace}
\newcommand{\scifive}{Sci5\xspace}
\newcommand{\gpt}{GPT-2\xspace}
\newcommand{\coder}{CODER\xspace}
\newcommand{\sapbert}{SapBERT\xspace}

\newcommand{\smm}{{SMM4H}\xspace}
\newcommand{\cadec}{{CADEC}\xspace}
\newcommand{\prop}{{PROP}\xspace}

\newcommand{\pretraining}{OP\xspace}
\newcommand{\finetuning}{FT\xspace}
\newcommand{\ptft}{OP+FT\xspace}

\newcommand{\seenmarker}{{\large$\circ$}\xspace}
\newcommand{\unseenmarker}{{\large$\diamond$}\xspace}
\newcommand{\overallmarker}{{\small$\bullet$}\xspace}

\newcommand{\anon}[1]{{[anon]}}
\newcommand{\argmax}{\mathop{\mathrm{argmax}}}

\newcommand{\std}[1]{{\color{gray}\footnotesize \textpm #1}}

\newcommand{\mypar}[1]{\noindent\textbf{#1}}

\newcommand{\seen}{\textsc{in}\xspace}
\newcommand{\unseen}{\textsc{out}\xspace}

\title{Generalizing over Long Tail Concepts for Medical Term Normalization}

\author{
Beatrice Portelli$^{1,2*}$
$\quad$
Simone Scaboro$^{1*}$
$\quad$
Enrico Santus$^{3\dag}$
\\
\textbf{Hooman Sedghamiz}$^3$ 
$\quad$
\textbf{Emmanuele Chersoni}$^4$
$\quad$
\textbf{Giuseppe Serra}$^1$
\\
{\small\shortstack{\\
$^1$ University of Udine, Italy \\
$^2$ University of Naples Federico II, Italy \\
$^3$ DSIG - Bayer Pharmaceuticals, New Jersey, USA\\
$^4$ The Hong Kong Polytechnic University, Hong Kong\\
\\
\texttt{ \{portelli.beatrice,scaboro.simone\}@spes.uniud.it,
esantus@gmail.com,}\\
\texttt{hooman.sedghamiz@bayer.com,
emmanuele.chersoni@polyu.edu.hk,
giuseppe.serra@uniud.it}
}}
}

\begin{document}
\maketitle

\def\thefootnote{*}\footnotetext{Equal contribution}
\def\thefootnote{\dag}\footnotetext{The author was affiliated with Bayer Pharmaceuticals at the time of the experiments, and is currently affiliated with Bloomberg.}
\def\thefootnote{\arabic{footnote}}

\begin{abstract}
Medical term normalization consists in mapping a piece of text to a large number of output classes.
Given the small size of the annotated datasets and the extremely long tail distribution of the concepts, it is of utmost importance to develop models that are capable to generalize to scarce or unseen concepts.
An important attribute of most target ontologies is their hierarchical structure. In this paper we introduce a simple and effective learning strategy that leverages such information to enhance the generalizability of both discriminative and generative models.
The evaluation shows that the proposed strategy produces state-of-the-art performance on seen concepts and consistent improvements on unseen ones, allowing also for efficient zero-shot knowledge transfer across text typologies and datasets.

\end{abstract}

\section{Introduction}

Term normalization is the task of mapping a variety of natural language expressions to specific concepts in a dictionary or an ontology. It is a key component for information processing systems, and it is extensively used in the medical domain. In this context, term normalization is often used to map reported adverse events (AEs) related to a drug to a medical ontology, such as \meddra \citep{meddra}. This is a challenging task, due to the high variability of natural language input (i.e., from the informality of social media and conversational transcripts to the formality of medical and legal reports) and the high cardinality and long tail distribution of the output concepts.
AEs are usually mappable to different levels of the same ontology: low-level concepts, which are closer to layman terms, and higher level concepts, which encompass the meaning of multiple low-level concepts. In \meddra,\footnote{{\meddra is a five-level hierarchy \url{https://www.meddra.org/how-to-use/basics/hierarchy}, but in this work we mainly focus on two of the levels: PT and LLT.}}
these two sets of concepts are called Lowest Level Terms (LLT), and Preferred Terms (PT) respectively; both of them have a very high cardinality (48,713 for LLT and 24,571 for PT, in \meddra version 23.1).
The following are examples of AEs, with their corresponding LLTs and PTs:\\

{\small
\begin{tabular}{ccc}
\textbf{AE} & \textbf{LLT} & \textbf{PT} \\
feel like crap & feeling unwell & malaise \\
weak knees & weakness & asthenia \\
zap me of all energy & loss of energy & asthenia \\
\end{tabular}\\
}

\begin{figure}[htbp]
\centering
\includegraphics[width=\linewidth, trim={.3cm .5cm .4cm .3cm}, clip]{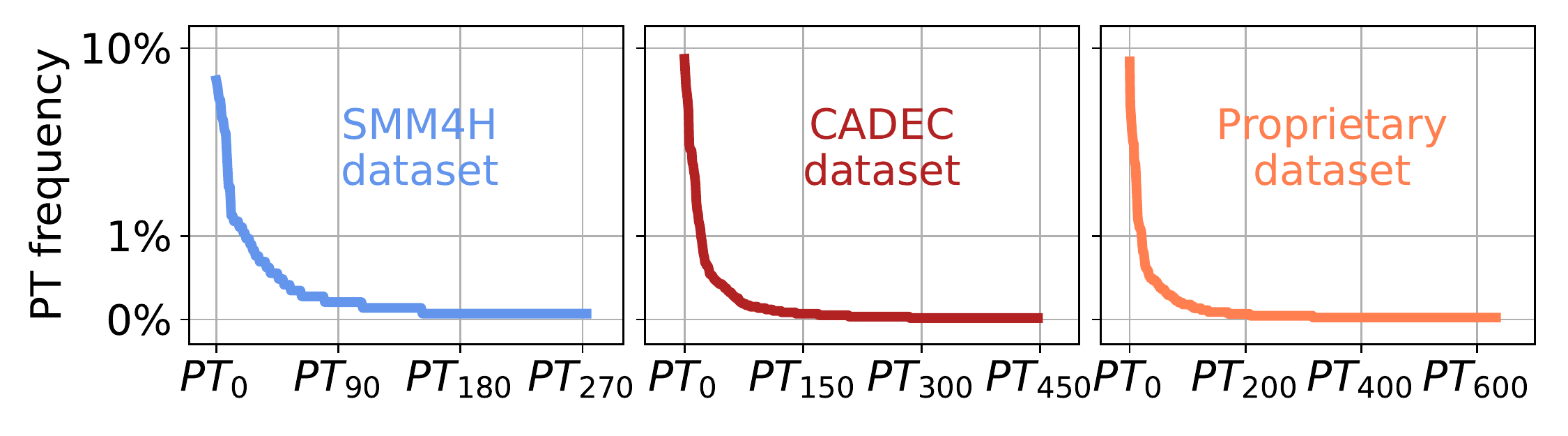}
\caption{Long tail distribution of PTs in the datasets used for this paper.}
\label{fig:long_tail}
\end{figure}

Currently this problem is addressed with large pretrained language models \citep{smm4h-2020-social}, finetuned on medical term normalization datasets, such as SMM4H \citep{smm4h-2020-social} or CADEC \citep{Karimi2015CadecAC}. 
However, these datasets contain maximum 5,000 samples, distributed on a few PT/LLT classes, and with a long tail distribution (see Figure \ref{fig:long_tail}).
Due to the size and distribution of these datasets, the resulting models usually perform well on examples that are seen in the training, but struggle to generalize on rare or unseen samples.

To improve the generalization capabilities of the models on long tail concepts, in this paper, we suggest to leverage the hierarchical nature of the medical ontology to enrich the large language models with domain knowledge before finetuning on a given training set.
Extensive experimental evaluation on three different datasets shows that the proposed strategy can be successfully applied to various model typologies, and it consistently outperforms other mainstream learning strategies, showing generalization capabilities not only across the long tail distribution, but also across text typologies and datasets.
The code and resources needed to replicate our experiments and test our learning strategy are publicly available\footnote{\url{https://github.com/AilabUdineGit/ontology-pretraining-code}}.

\section{Related Work}

Medical term normalization is generally regarded as either a classification or a ranking problem \citep{YUAN2022103983}. In the former case, a neural architecture encodes the term into a hidden representation and outputs a distribution over the classes \citep{limsopatham2016normalising,tutubalina2018medical,niu2019multi}, but this is difficult to scale to ontologies containing thousands of concepts, due to the absence of comprehensive datasets. 
In the ranking approach, on the other hand, the goal is to rank concepts by their similarity to the input term \citep{leaman2013dnorm,li2017cnn,sung2020biomedical}: a system is trained on binary classification, where terms and matching concepts are the positive samples, while terms and non-matching concepts are the negative ones. The raw output of the model is then used to rank the concepts.

Recent work successfully combined the two approaches. \citet{ziletti2022medical} presented a system mixing a BERT-based classifier \citep{devlin2019bert} and a zero/few-shot learning method to incorporate label semantics in the input instances \citep{halder2020task}, showing improved performance in single model and in ensemble settings.

Finally, systems like \coder \citep{YUAN2022103983} and SapBERT \citep{sapbert} introduced novel contrastive pretraining strategies that leverage UMLS to improve the medical embeddings of BERT-based models. While \sapbert leverages self-alignment methods, \coder maximizes the similarities between positive term-term pairs and term-relation-term triples and it claimed state-of-the-art results on several tasks, including zero-shot term normalization.
Another recent work by \citet{krissbert} introduced an even more extensive pretraining procedure, based on self-supervision and a combination of the traditional masked language modelling with contrastive losses . The strategy proved to be extremely effective for medical entity linking, a kind of term normalization which makes use of the full original context (instead of using only the AE).

\section{Proposed Learning Strategy: \ptft}

\newcommand{\centercol}[1]{\multicolumn{1}{c}{#1}}
\begin{table*}[!htbp]
\centering

\resizebox{.6\linewidth}{!}{
\begin{tabular}{cr rrr c}
\hline

\textbf{Dataset} &
\centercol{\textbf{\shortstack{Total\\Samples}}} &
\centercol{\textbf{\shortstack{Train\\Samples}}} &
\centercol{\textbf{\shortstack{Test\\Samples}}} &
\centercol{\textbf{\shortstack{\%\unseen\\Samples}}} &
\centercol{\textbf{\shortstack{Unique\\PTs}}}
\\
\hline
{\smm}   & 1,442 &  868 \std{06} &  574 \std{06} &
12.49 \std{0.94} & 274 \\
{\cadec} & 5,866 & 3,540 \std{21} & 2,326 \std{21} &
4.65 \std{0.41} & 488 \\
{\prop}  & 4,453 & 2,658 \std{65} & 1,796 \std{65} &
10.02 \std{0.79} & 634 \\
\hline
\end{tabular}
}
\caption{Dimensions of the datasets, reporting the average figures over the three train/test splits (\textpm\ std), as well as the number of unique PT terms contained in each dataset.}
\label{tab:dataset_size}
\end{table*}

Let's consider a target ontology (e.g. \meddra v23.1) containing two sets of concepts $\mathit{PT}=\{p_i\}$ and the $\mathit{LLT}=\{\ell_i\}$.
The ontology is structured so that every $\ell_i$ has only one parent $p_j$: $\mathit{parent}(\ell_i)=p_j$, but each $p_j$ can be parent of many $\ell_i$.
Given a set of Adverse Events $\mathit{AE}=\{a_i\}$, every $a_i$ can be univocally mapped to a $p_j$: $\mathit{norm}(a_i)=p_j$.

Our objective is to train a large language model $\mathcal{M}$ to encode $\mathit{norm}$: given a sample $(a_i, p_j)$, such that $\mathit{norm}(a_i)=p_j$, we want $\mathcal{M}(a_i)=p_j$.

\medskip

We propose a learning strategy based on the hierarchical structure of the ontology, composed of two steps: Ontology Pretraining and Finetuning.

During the first step, we expose the language model $\mathcal{M}$ to all possible output classes $p_j$ by leveraging the intrinsic hierarchical relation between LLT and PT. Specifically, we use the $\mathit{parent}$ relation to create a new set of training samples from the ontology, pairing each $\ell_i$ to its parent concept $p_j$.
In the case of \meddra, the new set of samples contains 48,713 $(\ell_i, p_j)$ pairs, where each $p_j$ appears multiple times, associated with different $\ell_i$. For example the PT ``asthenia'' will appear in the samples (weakness, asthenia), and (loss of energy, asthenia).
As LLTs are more informal than PTs, the language model $\mathcal{M}$ can be pretrained on this new set of data to gain general knowledge about all the target classes.
This pretraining set is highly similar to our target dataset ($\mathcal{M}(a_i)=p_j$), increasing the model transfer capability. We call this process ``Ontology Pretraining'' (\textbf{OP}).

The second step consists in finetuning (\textbf{FT}) an OP model on a specific term normalization dataset, which maps every AE $a_i$ to the corresponding PT $p_j$.
This step is crucial because the OP model lacks specific knowledge about the natural language style of real-world samples. Finetuning will also exploit the dataset sample distribution to boost the model's accuracy on the specific set of $p_j$ in the training set.
Note that the FT step can also be applied to a regular model without OP, resulting in a regular finetuning.

We hypothesize that the combination of \textbf{OP+FT} to a discriminative or a generative language model $\mathcal{M}$ will improve its performance on seen concepts in the training set, while making it more generalizable to long tail and unseen concepts.

\section{Experimental Setting}

\subsection{Datasets}

To investigate the performance of our learning strategy, we used three English datasets for \meddra term normalization, with different writing styles.

\mypar{\smm} \citep{smm4h-2020-social}.
Public dataset for the challenge SMM4H 2020 - Task 3, AE normalization. It contains 2,367 tweets, 1,212 of which report AEs with highly informal language, mapped to a PT/LLT.

\mypar{\cadec} \citep{Karimi2015CadecAC}.
Public dataset containing 1,250 posts from the health forum ``AskaPatient'', containing user-reported AEs mapped to a PT/LLT. The language is informal, but still more medically precise than \smm.

\mypar{\prop}.
Proprietary dataset provided by Bayer Pharmaceuticals, containing 2,010 transcripts of phone-calls with healthcare professionals reporting their patients' AEs, mapped to PTs. The language is more formal and medically accurate.

\subsection{Data Preparation}

All datasets were preprocessed to obtain samples containing only $(a_i,p_j)$ pairs. The samples in \cadec and \smm which were labelled with an $\ell_i \in \mathit{LLT}$ were re-labelled with $\mathit{parent}(\ell_i)=p_j \in \mathit{PT}$, obtaining a uniform output space for all datasets containing only PT concepts.

Since the focus of this work is on the generalization capabilities of the models, it is important to test the models on different sets of unseen labels. 
For this reason, we created three random splits of train/test samples using a 60:40 proportion, instead of using the public fixed train/test split.
Given a train and a test set, every test sample with label $p_j$ falls into one of the following categories:\\
-- \seen, if $p_j$ is present in the training set;\\
-- \unseen, if $p_j$ is \textit{not} present in the training set.

\noindent The most important set of samples to measure the generalization capabilities of the models is \unseen.

Table \ref{tab:dataset_size} reports figures for the resulting datasets. \cadec and \prop contain the largest number of samples (5,866 and 4,453 respectively), while \smm is sensibly smaller, with only 1,442 samples.
The largest datasets also contain the largest number of PTs: 488 for \cadec and 634 for \prop. \smm only contains 274 PTs instead. Most of the PTs are unique to one of the three datasets and do not appear in the other ones, making it impossible to gain a substantial advantage by combining them (see Appendix \ref{app:venn}).
We observe that the percentage of \unseen samples varies from 5\% to 12\%, with \smm being the most challenging dataset. The standard deviation is low, showing that the presence of 5--12\% \unseen samples is a characteristic of the specific dataset, resulting from its long tail PT distribution. Note also that the smaller the dataset, the higher the percentage of \unseen samples in the test set.

\subsection{Models}

To test the proposed strategy and observe how it affects generalization, we selected different kinds of widely-adopted models.
In particular, we compare PubMedBERT \citep{pubmedbert}, \scifive \citep{https://doi.org/10.48550/arxiv.2106.03598}, \gpt \citep{radford2019language}, \coder \citep{YUAN2022103983} and \sapbert \citep{sapbert}.

\mypar{PubMedBERT (\pubmedbert).} It was chosen as an example of a BERT-based classifier due to its medical pretraining (PubMed articles) and strong performance in other medical tasks \citep{pubmedbert,portelli2021bert,scaboro2021nade,scaboro2022increasing}.

\mypar{\gpt and \scifive.} \gpt was selected as an example of a general-purpose autoregressive language model for text generation, while \scifive was chosen for its medical pretraining, performed on the same kind of texts as \pubmedbert.
The models were trained to generate a PT, given an input prompt containing the adverse event.

\mypar{\coder and \sapbert (SapB).}
To the best of our knowledge, \coder and \sapbert are some of the best dataset-agnostic models for medical term embeddings. They were both trained on the UMLS ontology \citep{umls}, which is a super-set of \meddra, and tested on several term normalization datasets, showing promising results.
Following both original papers, we use \coder and \sapbert to generate embeddings for $a_i$ and for all $p_j \in \mathit{PT}$. We then select as prediction the $p_j$ that minimizes the cosine similarity with $a_i$.

We also trained both models according to our proposed strategy. Both models were trained using the contrastive settings described in their paper and the respective codebases\footnote{\coder: \url{https://github.com/GanjinZero/CODER}\\ \sapbert: \url{https://github.com/cambridgeltl/sapbert} }.

See Appendix \ref{app:training_details} for training details for all models and \ref{app:coder_details} for more details on the contrastive training of \coder and \sapbert.

Performance is assessed with the Accuracy metric, but we also report the F1 Score in Appendix \ref{app:full_results}, as it can give more insights when classes are unbalanced.

\section{Experimental Results}

\begin{figure}[!htbp]
\small\centering

\includegraphics[width=.93\linewidth, trim={ 0cm   0.4cm 0.2cm 0.5cm}, clip]{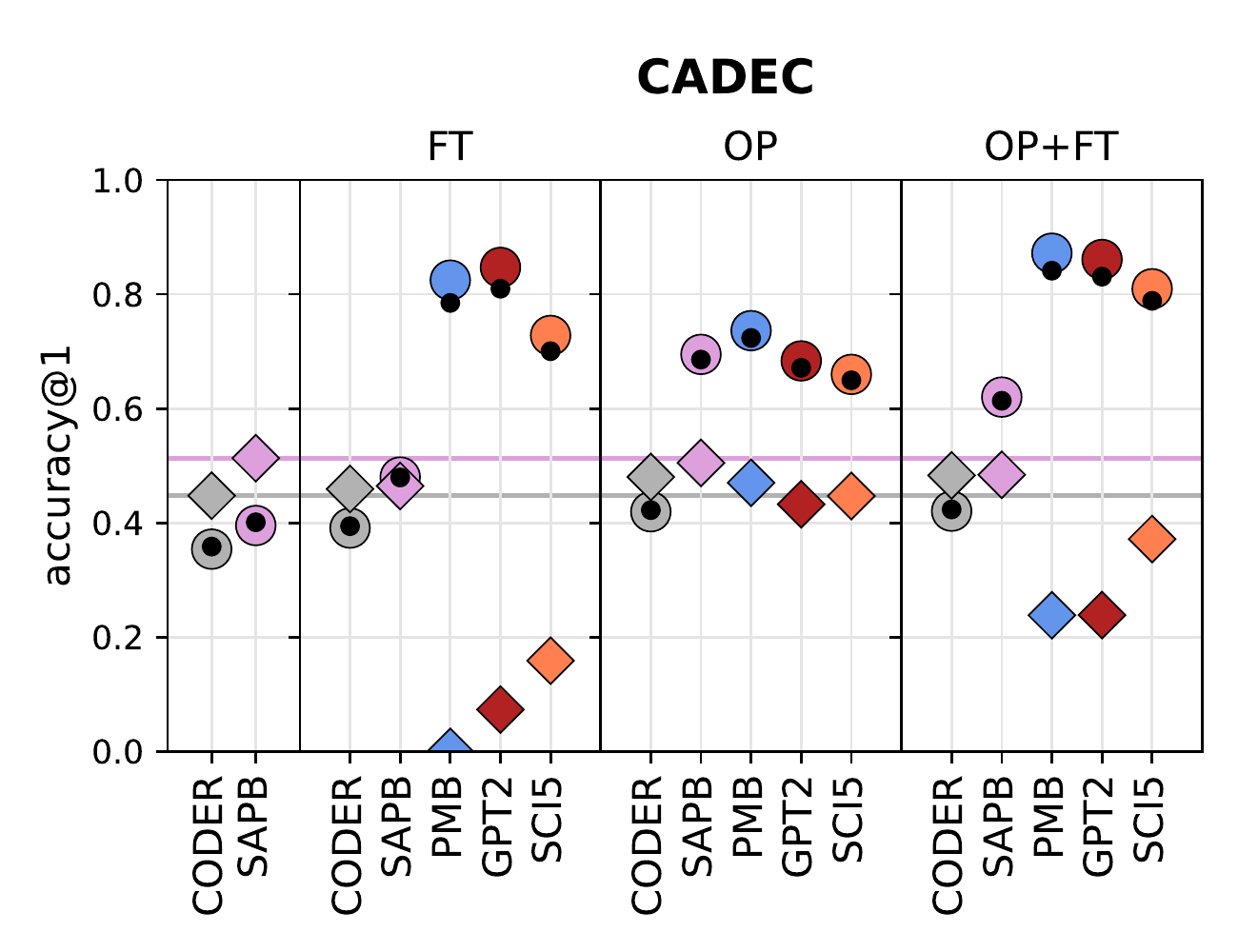}\\ \medskip
\includegraphics[width=.93\linewidth, trim={ 0cm   0.4cm 0.2cm 0.5cm}, clip]{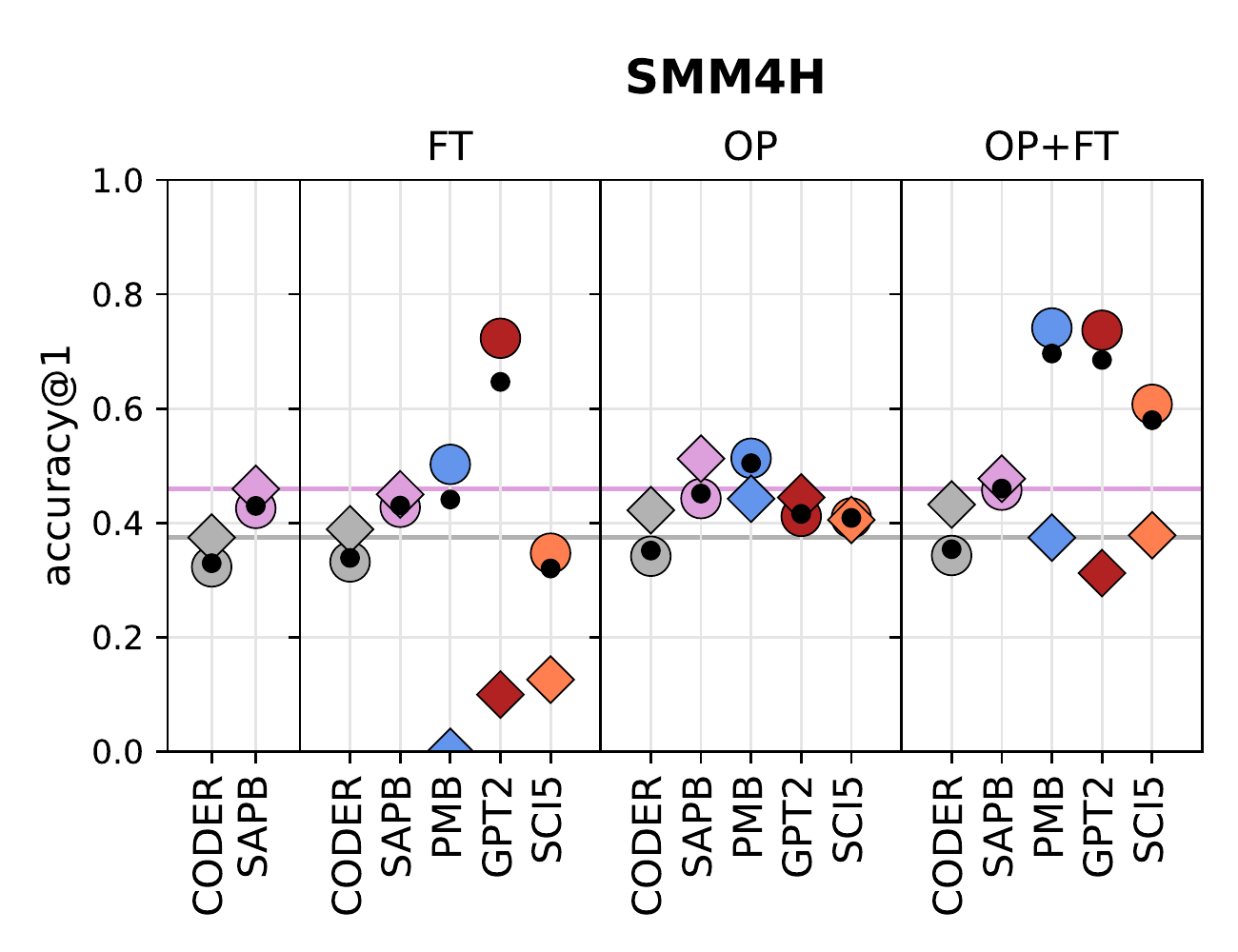}\\ \medskip
\includegraphics[width=.93\linewidth, trim={ 0cm   0.4cm 0.2cm 0.5cm}, clip]{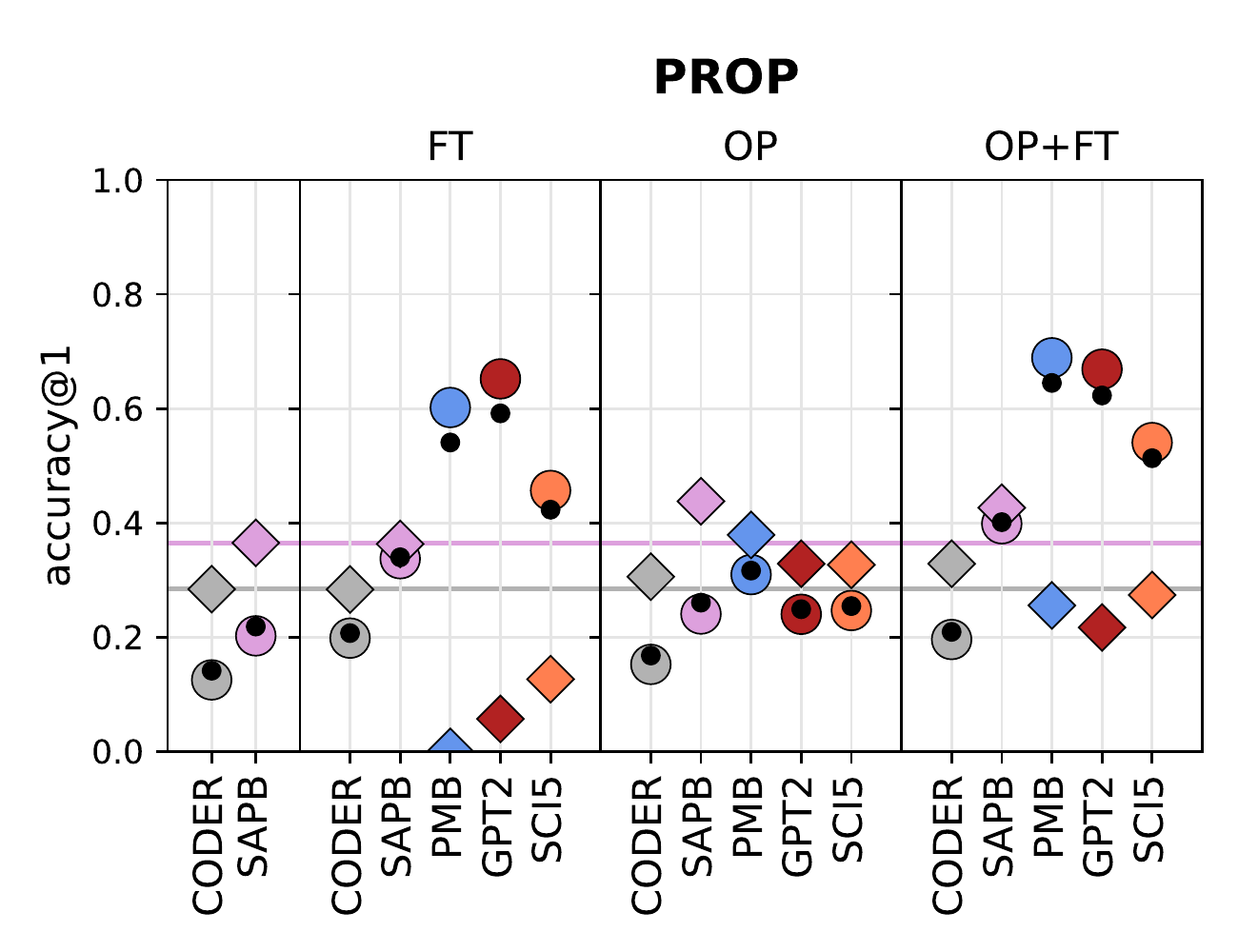} 

\caption{Accuracy of all models on the three datasets on \seen (\seenmarker), \unseen (\unseenmarker) and all (\overallmarker) samples.}
\label{fig:all_datasets}
\end{figure}

In an ablation-study fashion, we compare the \ptft learning strategy with its two components: \pretraining and \finetuning.
Figure \ref{fig:all_datasets} contains the results for all the tested models and training strategies, and is organized as follows. We display a plot for each dataset, reporting the accuracy of the models on \seen samples (\seenmarker), \unseen samples (\unseenmarker) and the whole test set (\overallmarker).
The first column shows the performance of a basic \coder and \sapbert model without any additional training. We consider their accuracy on \unseen (\unseenmarker) as our generalization goal, and plot them as solid lines across the chart. The following three columns display the performance of all the models, trained with one of the learning strategies (FT, OP and OP+FT respectively). For tabular results, see Appendix \ref{app:full_results}.

\textbf{\coder} and \textbf{\sapbert} on their own proved to be strong baselines across the three datasets.
Looking at the first column, they reach 40--50\% accuracy on \cadec and \smm (overall, \seen and \unseen, see solid lines), and around 15--20\% overall accuracy (\overallmarker) on \prop.

All learning strategies seemed to be ineffective on \coder: its performance (gray markers) remains roughly the same across all strategies (FT, OP or OP+FT).
A possible explanation for this behaviour is that \coder embeddings are already in an optimal state according to the training objectives, as they have been trained on a very similar task. In fact, \coder generates predictions using the similarity between the embeddings, and the stable performance indicates that there were no drastic changes in the structure of the embedding space.

A clearer effect of the training strategies can be seen on \sapbert (lilac markers), although it is still limited when compared with the other models. \sapbert embeddings are probably more subject to adjustments compared to \coder because the latter was trained for significantly more steps and using more objective functions, leading to less-adaptable embeddings.

The following observations apply the other three models: \pubmedbert, \gpt and \scifive.

The \textbf{\finetuning} strategy (second column), as expected, works really well for \seen (\seenmarker) samples: on \cadec the \seen\ accuracy of all models is over to 80\%, while it is close to 50\% for the other two datasets.
However, the \unseen\ accuracy (\unseenmarker) is lower than 20\% in all cases (significantly lower than the solid line), and reaches 0\% for \pubmedbert, showing that finetuning alone is not sufficient for classifiers to generalize on \unseen samples in this setting.

The \textbf{OP} strategy (third column), brings the \unseen accuracy of all models on pair with the \coder/\sapbert baselines, while the \seen/overall accuracy matches or surpasses them. Comparing OP with FT, we see that the overall accuracy (\overallmarker) of the model is generally lower for OP. However, the performance on \unseen (\unseenmarker) samples doubles for generative models, and jumps from 0 to ~40\% for \pubmedbert.
This shows that the first step of our proposed learning strategy has the desired effect, as it improves the models' understanding of all the output classes.

Finally, looking at the models trained with the \textbf{\ptft} strategy (fourth column), we see that they outperform the FT ones on overall and \seen\ accuracy. The effect is particularly strong on the \smm dataset (cf. \pubmedbert FT, 44\% and \pubmedbert OP+FT, 70\%).
At the same time, the performance on \unseen (\unseenmarker) samples remains similar to the \pretraining models and close to the \coder baseline (gray solid line). The only exception is \cadec, where the performance on \unseen is in-between the baseline and the accuracy with \finetuning only.
This shows that the proposed \ptft learning strategy can successfully improve the generalization capabilities of various language models, while also improving their overall performance.

\begin{figure}[htbp]
\small\centering
\includegraphics[height=3.7cm, trim={0.2cm .4cm .3cm .2cm}, clip]{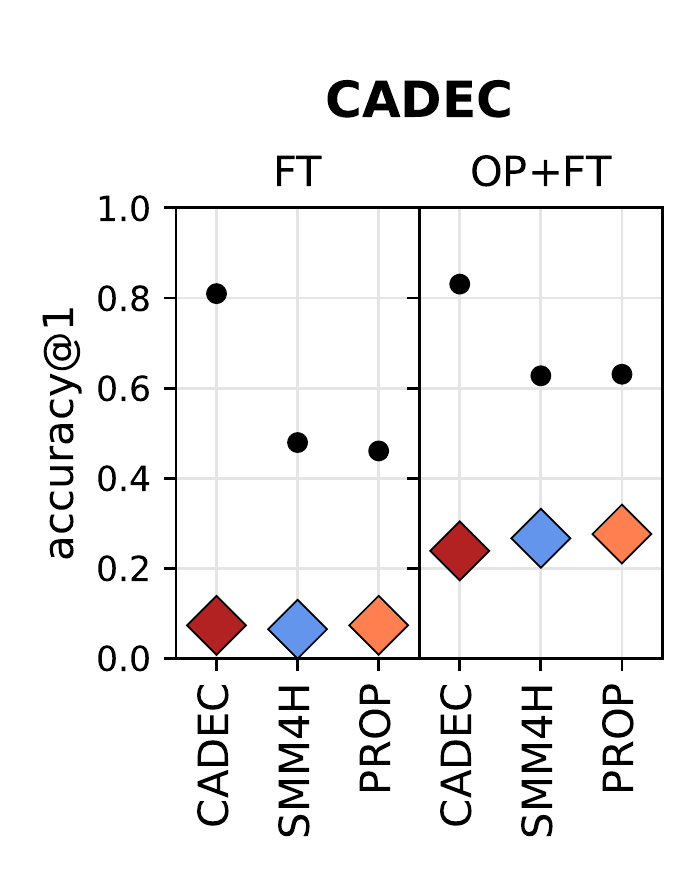}
\includegraphics[height=3.7cm, trim={1.7cm .4cm .3cm .2cm}, clip]{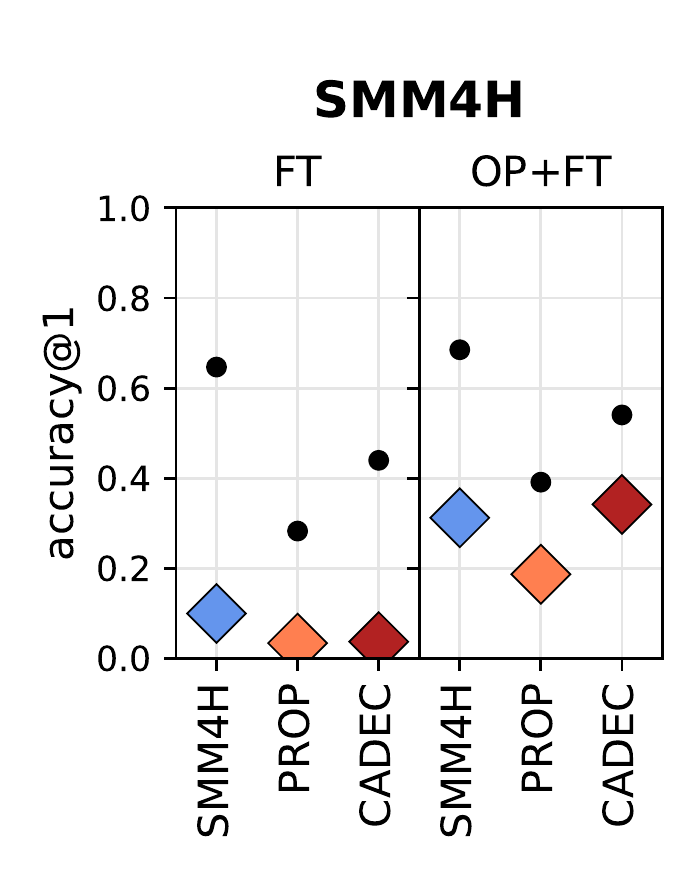}
\includegraphics[height=3.7cm, trim={1.7cm .4cm .3cm .2cm}, clip]{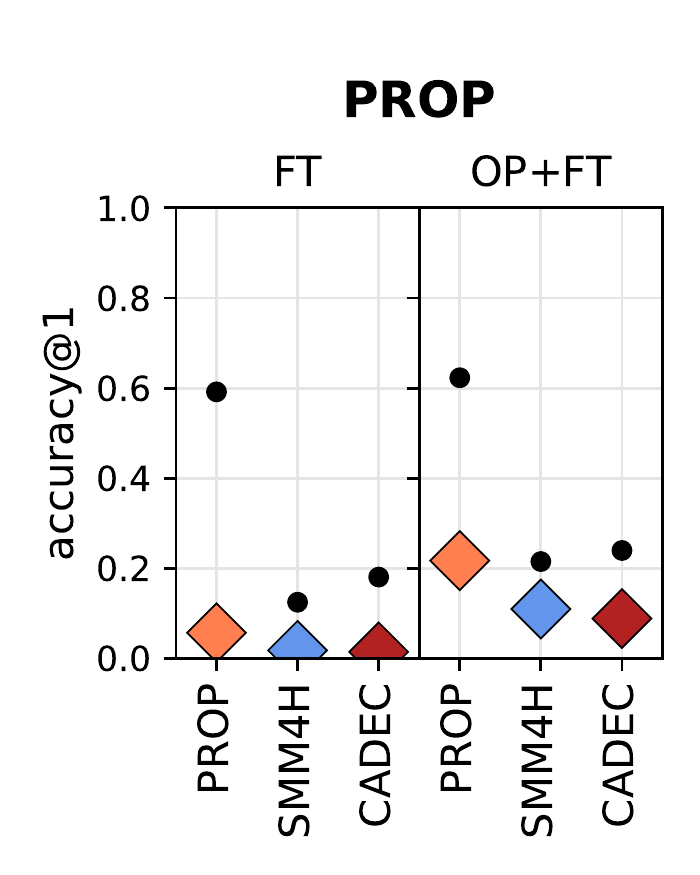}\\

\caption{Cross-dataset accuracy for \gpt (\finetuning and \ptft) on \unseen (\unseenmarker) and all (\overallmarker) samples. One plot for each \textit{test} dataset; the x-axis reports the \textit{training} dataset.}\label{fig:all_datasets_transfer}
\end{figure}

We further test the generalization of \ptft models in zero-shot, cross-dataset term normalization, normalizing the terms of each dataset with models that have been trained on one of the other two. Figure \ref{fig:all_datasets_transfer} shows the accuracy of \gpt, with a plot for each test dataset, different training datasets on the x-axis, and one column for each learning strategy (FT or \ptft). The behaviour of the other models is similar (see Appendix \ref{app:full_results}).
In all columns, we observe a drop in overall accuracy (\overallmarker) between the first data point  and the following ones (e.g., cf. \cadec trained on \cadec and \cadec trained on \smm). However, this drop is larger for FT models than \ptft ones (e.g., 35 vs. 20 points on \cadec). In addition, the \unseen\ accuracy of \ptft models remains high regardless of the training set. This shows that \ptft models generalize better than FT models across-dataset.
Note that generalization is still challenging when moving from a dataset with highly informal language to a formal one (see \prop trained on \smm).

\section{Conclusions}
In this paper, we shed some light on the importance of generalization for medical term normalization models. We showed that AE normalization models trained with traditional finetuning, despite showing high accuracy on leaderboards, have low generalization capabilities due to the long tail distribution of the target labels.
Our proposed training strategy (\ptft), which leverages the hierarchical structure of the ontology, outperforms traditional models, while also obtaining state-of-the-art results in generalization on \unseen samples. This was also demonstrated in a zero-shot normalization setting.
\ptft showed improvements on discriminative and generative language models, while it seems to be less effective on models trained with contrastive losses.
This promising technique could also be applied to other tasks with massive output spaces organized in a hierarchical manner.

\section*{Limitations}

The proposed learning strategy was tested only for the task of medical term normalization (from adverse events to \meddra concepts). However, it would be interesting to test its effectiveness on other term normalization tasks beyond \meddra mapping and outside of the medical domain.

Even restricting the problem to medical term normalization, and using datasets with different text styles, we only focused on English texts. Medical ontologies such as \meddra and UMLS are released in multiple languages, and the research community is moving towards multi-lingual approaches. In the future, we plan to extend this strategy to other languages (such as Spanish and Chinese) and to test the models' capacity to perform crosslingual transfer in zero-shot scenarios.

\section*{Acknowledgements}
The authors thank Juergen Dietrich, Senior Lead Data Scientist at Bayer Pharmaceuticals, for the help in the creation and annotation of the \prop dataset. Thanks also to the three anonymous reviewers for their insightful comments.

\bibliography{custom}
\bibliographystyle{acl_natbib}

\appendix

\section{Training Specifications \label{app:training_details}}

Table \ref{tab:training_specs} contains the specifics for the number of Ontology Pretraining (OP) and Finetuning (FT) used for all the selected models.

\begin{table}[!htbp]
\centering
\small
\begin{tabular}{cccc@{\ +}c}
\hline
\textbf{Model} &
\textbf{\shortstack{\\\pretraining \\ epochs}} &
\textbf{\shortstack{\\\finetuning \\ epochs}} &
\multicolumn{2}{c}{\textbf{\shortstack{\\\ptft\\ epochs}}} \\
\hline
\pubmedbert &  30 & 10 & 30 &  5 \\
\gpt        &  30 & 15 & 30 & 10 \\
\scifive    &  40 & 15 & 40 &  8 \\
\coder      &  50 & 20 & 50 & 15 \\
\sapbert    &  30 & 15 & 30 & 10 \\
\hline
\end{tabular}
\caption{Further details about training parameters}
\label{tab:training_specs}
\end{table}

Other model-related details:

\begin{itemize}
\item \textbf{\pubmedbert} A classification head (24,571 output classes) was added to the base model.
\item \textbf{\gpt} Given a sample $(a_i, p_j)$, the input prompt for the model was \texttt{{\sf"}INPUT: $a$\textbackslash nMEANING:{\sf"}}. The model was trained to complete the sentence with $p_j$.
\item \textbf{\scifive} Given a sample $(a_i, p_j)$, the input prompt for the model was \texttt{{\sf"}normalize: $a${\sf"}}. The model was trained to respond with a string containing $p_j$.

\item \textbf{\coder\xspace/\xspace\sapbert} Following their original papers, we use \coder/\sapbert to normalize an AE $a$ as follows:
$$\hat{p} = \argmax_{p \in P} \mathit{sim}(\mathcal{C}(a),\mathcal{C}(p))$$
\noindent where $\mathit{sim}$ is a similarity measure (cosine, in our case), and $\mathcal{C}(\cdot)$ is the result of embedding a term with \coder/\sapbert. $\hat{p}$ is the predicted PT, which is compared with the actual one to evaluate the model.
\end{itemize}

\section{Sample Creation for Contrastive Training \label{app:coder_details}}

\subsection{\coder}

\coder leverages on term-term pairs and term-relation-term triples for its contrastive training strategy.
We create positive/negative samples for the term-term pairs using the AEs having equal/different PT, and term-relation-term triples connecting AEs whose PTs have the same $\mathit{parent}$.

For example, let's consider the following $(a_i, p_j)$ samples, for which we also report $\mathit{parent}(p_j)$:\\

\smallskip

{\small
\begin{tabular}{ccc}
\hline
{$a_i$} & {$p_j$} & {$\mathit{parent}(p_j)$} \\
\hline
feel like crap       & malaise  & Asthenic conditions	\\
weak knees           & asthenia & Asthenic conditions	\\
zap me of all energy & asthenia & Asthenic conditions	\\
\hline
\end{tabular}\\
}

\medskip

This will generate the following training samples for \coder:

\begin{itemize}
\item \textbf{positive term-term:}\\(weak knees, zap me of all energy)\\because they share the same $p_j$ ``asthenia''
\item \textbf{negative term-term:}\\(weak knees, feel like crap) and\\(zap me of all energy, feel like crap) \\because they are labelled with a different $p_j$ (``asthenia'' vs ``malaise'')
\item \textbf{positive term-relation-term}:\\(weak knees, RO, feel like crap) and\\(zap me of all energy, RO, feel like crap),\\because their $p_j$ share the same $\mathit{parent}$ ``Asthenic conditions''. RO stands for ``Related Other'', one of the standard term relations defined in the UMLS ontology, and we use it to encode the relation ``same granparent''.
\end{itemize}

This sample generation procedure is repeated for all samples in the three datasets (\smm, \cadec and \prop), as well as for the additional samples generated from \meddra for the OP strategy.

\subsection{\sapbert}

\sapbert leverages on term-term synonym pairs, where the positive pairs belong to the same upper-level concept.

The finetuning script present in the GitHub repository requires a list of term pairs belonging to the same concept.
In the the case of the three datasets (\smm, \cadec and \prop) we generate the terms pairs as $(\ell_i, a_j)$, where $\mathit{parent}(\ell_i) = \mathit{norm}(a_j)$.
For the OP strategy, the samples are all possible pairs $(\ell_i, \ell_j)$, where $\mathit{parent}(\ell_i) = \mathit{parent}(\ell_j)$.

\section{Dataset Comparison \label{app:venn}}

\begin{figure}[!htbp]
    \centering
    \includegraphics[width=.7\linewidth, trim={2cm 1.2cm 2cm 1.2cm}, clip]{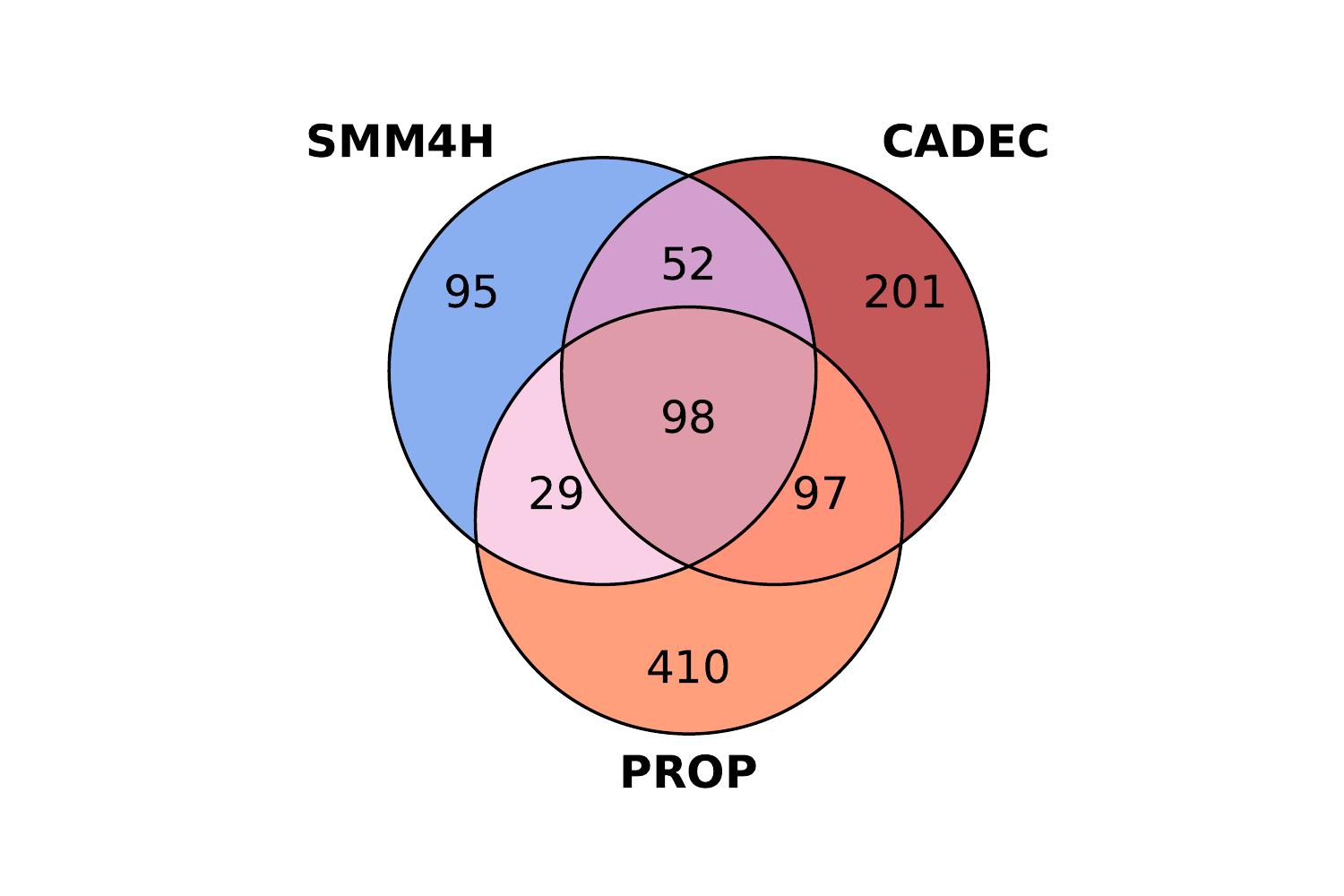}
    \caption{Venn diagram of the shared/unique PT concepts for the three datasets.}
    \label{fig:venn}
\end{figure}

Most of the PTs present in the three datasets are unique to a specific dataset, making it really challenging to perform transfer learning from one to the other without dealing with long-tail and unseen concepts.
The Venn diagram in Figure \ref{fig:venn} shows the number of PT concepts in common between all the datasets.
706 PTs are unique to one of the three datasets, 276 are shared among at least two datasets, and only 98 appear in all three of them.
Out of all the PTs in \prop, 64\% are unique (410 out of 634) to this dataset alone, making it the most challenging to perform cross-dataset normalization on. The following most challenging datasets are \cadec (41\% unique PTs) and \smm (34\% unique PTs).

\section{Complete Results \label{app:full_results}}

Tables \ref{tab:full_results_smm}, \ref{tab:full_results_cadec} and \ref{tab:full_results_prop} include the full results of all tested models (both Accuracy and F1 Score).

Tables \ref{tab:cross_coder}, \ref{tab:cross_sapbert}, \ref{tab:cross_pmb}, \ref{tab:cross_gpt} and \ref{tab:cross_scifive} report the accuracy for all the cross-dataset experiments (one table for each model).

\begin{table*}[!htbp]
\center

\textbf{\large\smm}\\ \medskip

\resizebox{.9\linewidth}{!}{
\begin{tabular}{ll ccc  }
\hline
\multicolumn{2}{c}{} &
\multicolumn{3}{c}{\textbf{Accuracy}} \\
 &&  \textbf{\seen} &  \textbf{\unseen} &  \textbf{overall}  \\
\hline \multirow{2}{*}{\rotatebox{90}{}}
& CODER   & 32.34 \std{1.22} &  37.47 \std{2.07} &  32.98 \std{1.24} \\
& SAPB    & 42.56 \std{0.50} &  45.98 \std{3.60} &  43.00 \std{0.77} \\
\hline \multirow{5}{*}{\rotatebox{90}{\finetuning}}
& CODER   &  33.20 \std{1.53} &  38.91 \std{1.97} &  33.90 \std{1.58} \\
& SAPB    &  42.76 \std{1.58} &  45.02 \std{4.84} &  43.05 \std{1.96} \\
& PMB     &  50.26 \std{0.95} &  00.00 \std{0.00} &  44.12 \std{1.24} \\
& GPT2    &  72.31 \std{1.55} &  10.02 \std{2.21} &  64.68 \std{0.98} \\
& SCI5    &  34.73 \std{2.05} &  12.63 \std{3.33} &  32.06 \std{1.84} \\
\hline \multirow{5}{*}{\rotatebox{90}{\pretraining}}
& CODER   &  34.20 \std{0.62} &  42.25 \std{1.86} &  35.19 \std{0.65} \\
& SAPB    &  44.29 \std{1.91} &  51.25 \std{6.04} &  45.15 \std{2.31} \\
& PMB     &  51.33 \std{1.15} &  44.25 \std{2.10} &  50.46 \std{1.18} \\
& GPT2    &  41.17 \std{0.52} &  44.48 \std{4.90} &  41.60 \std{1.05} \\
& SCI5    &  40.90 \std{0.80} &  40.56 \std{6.90} &  40.90 \std{1.48} \\
\hline \multirow{5}{*}{\rotatebox{90}{\ptft}}
& CODER   &  34.33 \std{0.80} &  43.24 \std{1.40} &  35.43 \std{0.83} \\
& SAPB    &  45.75 \std{0.66} &  47.76 \std{7.09} &  46.03 \std{1.36} \\
& PMB     &  74.10 \std{2.55} &  37.46 \std{4.10} &  69.64 \std{1.80} \\
& GPT2    &  73.71 \std{1.29} &  31.25 \std{3.37} &  68.52 \std{1.30} \\
& SCI5    &  60.76 \std{0.91} &  37.87 \std{3.21} &  57.98 \std{0.97} \\
\hline
\end{tabular}\hspace{-.1cm} %
\begin{tabular}{|ccc}
\hline
\multicolumn{3}{c}{\textbf{F1 Score}} \\
\multicolumn{1}{c}{\textbf{\seen}} &  \textbf{\unseen} &  \textbf{overall}  \\
\hline
13.52 \std{0.96} &  23.67 \std{1.42} &  16.34 \std{1.14} \\
16.91 \std{0.57} &  29.93 \std{3.32} &  20.23 \std{0.20} \\ \hline
14.12 \std{1.22} &  24.67 \std{1.16} &  17.14 \std{1.34} \\
16.80 \std{0.78} &  29.58 \std{4.68} &  20.10 \std{1.17} \\
21.43 \std{1.91} &  00.00 \std{0.00} &  13.68 \std{1.43} \\
11.51 \std{0.49} &  07.09 \std{1.68} &  10.43 \std{0.48} \\
13.77 \std{0.93} &  05.27 \std{2.61} &  11.83 \std{0.42} \\ \hline
14.25 \std{0.81} &  27.30 \std{1.20} &  17.93 \std{0.83} \\
17.96 \std{1.03} &  35.12 \std{4.84} &  22.27 \std{1.45} \\
25.87 \std{1.21} &  27.95 \std{0.84} &  27.51 \std{1.41} \\
17.50 \std{0.89} &  24.29 \std{4.18} &  19.63 \std{1.47} \\
17.13 \std{1.05} &  22.70 \std{2.60} &  18.99 \std{1.37} \\ \hline
14.40 \std{0.64} &  28.19 \std{1.02} &  18.18 \std{0.65} \\
18.43 \std{1.17} &  32.04 \std{5.22} &  21.92 \std{1.66} \\
55.59 \std{3.08} &  22.49 \std{2.73} &  45.62 \std{2.51} \\
52.52 \std{2.61} &  19.59 \std{2.53} &  42.20 \std{2.68} \\
32.25 \std{2.85} &  23.50 \std{2.83} &  30.08 \std{1.77} \\ \hline
\end{tabular}
}

\caption{Full metrics (accuracy and F1 score) of all tested models on the \smm dataset. \label{tab:full_results_smm}}
\end{table*}
\begin{table*}[htbp]
\center

\textbf{\large\cadec}\\ \medskip

\resizebox{.9\linewidth}{!}{
\begin{tabular}{ll ccc }
\hline
{} &&
\multicolumn{3}{c}{\textbf{Accuracy}} \\
{} &&  \textbf{\seen} &  \textbf{\unseen} &  \textbf{overall} \\
\hline \multirow{2}{*}{\rotatebox{90}{}}
& CODER  &  35.44 \std{0.40} &  44.79 \std{3.35} &  35.89 \std{0.42} \\
& SAPB   &  39.57 \std{1.05} &  51.34 \std{4.67} &  40.14 \std{1.14} \\
\hline \multirow{5}{*}{\rotatebox{90}{\finetuning}}
& CODER   &  39.13 \std{1.54} &  45.96 \std{2.15} &  39.46 \std{1.45} \\
& SAPB    &  48.04 \std{4.03} &  46.48 \std{4.49} &  47.97 \std{3.70} \\
& PMB     &  82.47 \std{0.27} &  00.00 \std{0.00} &  78.49 \std{0.10} \\
& GPT2    &  84.70 \std{0.42} &  07.40 \std{0.13} &  80.97 \std{0.45} \\
& SCI5    &  72.80 \std{0.12} &  15.94 \std{1.37} &  70.05 \std{0.17} \\
\hline \multirow{5}{*}{\rotatebox{90}{\pretraining}}
& CODER   &  41.98 \std{0.65} &  48.06 \std{3.12} &  42.26 \std{0.65} \\
& SAPB    &  69.49 \std{0.37} &  50.53 \std{1.40} &  68.58 \std{0.37} \\
& PMB     &  73.64 \std{0.35} &  47.05 \std{0.50} &  72.36 \std{0.40} \\
& GPT2    &  68.35 \std{0.22} &  43.29 \std{2.47} &  67.14 \std{0.22} \\
& SCI5    &  65.99 \std{0.58} &  44.73 \std{2.22} &  64.96 \std{0.45} \\
\hline \multirow{5}{*}{\rotatebox{90}{\ptft}}
& CODER   &  42.08 \std{0.80} &  48.34 \std{2.86} &  42.38 \std{0.75} \\
& SAPB    &  62.02 \std{2.06} &  48.44 \std{1.74} &  61.38 \std{1.91} \\
& PMB     &  87.18 \std{0.33} &  23.88 \std{1.79} &  84.12 \std{0.28} \\
& GPT2    &  86.08 \std{0.78} &  23.92 \std{1.09} &  83.08 \std{0.69} \\
& SCI5    &  80.96 \std{0.40} &  37.21 \std{4.32} &  78.85 \std{0.52} \\
\hline
\end{tabular}\hspace{-.1cm} %
\begin{tabular}{|ccc}
\hline
\multicolumn{3}{c}{\textbf{F1 Score}} \\
\multicolumn{1}{c}{\textbf{\seen}} &  \textbf{\unseen} &  \textbf{overall}  \\
\hline
17.78 \std{0.43} &  26.81 \std{2.20} &  19.89 \std{0.58} \\
17.44 \std{0.63} &  31.62 \std{3.29} &  20.20 \std{0.56} \\ \hline
18.72 \std{0.54} &  27.53 \std{1.46} &  20.88 \std{0.38} \\
20.25 \std{1.50} &  27.02 \std{3.67} &  22.11 \std{0.62} \\
47.64 \std{1.29} &  00.00 \std{0.00} &  33.32 \std{1.44} \\
30.97 \std{0.50} &  08.31 \std{0.83} &  25.00 \std{0.40} \\
37.02 \std{1.63} &  07.11 \std{3.29} &  29.84 \std{1.39} \\ \hline
24.40 \std{1.78} &  27.12 \std{3.61} &  25.25 \std{1.95} \\
27.48 \std{0.81} &  32.38 \std{1.54} &  29.39 \std{0.32} \\
38.46 \std{1.36} &  32.21 \std{1.40} &  38.47 \std{0.65} \\
28.76 \std{1.20} &  28.75 \std{1.23} &  29.66 \std{1.35} \\
27.10 \std{1.09} &  27.27 \std{1.49} &  27.98 \std{0.86} \\ \hline
20.77 \std{0.66} &  29.52 \std{1.66} &  23.31 \std{0.71} \\
27.20 \std{0.82} &  29.45 \std{1.59} &  28.68 \std{0.54} \\
70.43 \std{1.00} &  15.74 \std{1.15} &  55.78 \std{0.36} \\
64.17 \std{2.89} &  15.07 \std{0.67} &  51.05 \std{1.04} \\
50.09 \std{2.36} &  22.66 \std{2.23} &  44.36 \std{2.11} \\ \hline
\end{tabular}
}

\caption{Full metrics (accuracy and F1 score) of all tested models on the \cadec dataset. \label{tab:full_results_cadec}}
\end{table*}

\begin{table*}[htbp]
\center

\textbf{\large\prop}\\ \medskip

\resizebox{.9\linewidth}{!}{
\begin{tabular}{ll ccc  }
\hline
{} &&
\multicolumn{3}{c}{\textbf{Accuracy}} \\
{} &&   \textbf{\seen} &  \textbf{\unseen} &  \textbf{overall} \\
\hline \multirow{2}{*}{\rotatebox{90}{}}
& CODER   & 12.58 \std{0.44} &  28.44 \std{3.02} &  14.19 \std{0.71} \\
& SAPB    & 20.28 \std{0.59} &  36.54 \std{3.61} &  21.91 \std{0.81} \\
\hline \multirow{5}{*}{\rotatebox{90}{\finetuning}}
& CODER   &  19.88 \std{0.89} &  28.39 \std{3.81} &  20.76 \std{1.20} \\
& SAPB    &  33.77 \std{2.57} &  36.34 \std{3.71} &  34.02 \std{2.63} \\
& PMB     &  60.20 \std{0.34} &  00.00 \std{0.00} &  54.09 \std{0.76} \\
& GPT2    &  65.21 \std{0.73} &  05.75 \std{1.22} &  59.18 \std{1.12} \\
& SCI5    &  45.70 \std{0.77} &  12.71 \std{1.13} &  42.34 \std{0.61} \\
\hline \multirow{5}{*}{\rotatebox{90}{\pretraining}}
& CODER   &  15.29 \std{0.24} &  30.62 \std{1.97} &  16.84 \std{0.38} \\
& SAPB    &  24.10 \std{0.76} &  43.80 \std{2.54} &  26.09 \std{0.85} \\
& PMB     &  30.99 \std{1.00} &  37.92 \std{1.66} &  31.69 \std{1.04} \\
& GPT2    &  24.05 \std{0.77} &  32.89 \std{2.23} &  24.94 \std{0.86} \\
& SCI5    &  24.71 \std{0.78} &  32.71 \std{4.80} &  25.49 \std{1.12} \\
\hline \multirow{5}{*}{\rotatebox{90}{\ptft}}
& CODER   &  19.64 \std{0.25} &  32.88 \std{2.51} &  20.98 \std{0.28} \\
& SAPB    &  39.88 \std{3.20} &  42.68 \std{3.66} &  40.16 \std{3.23} \\
& PMB     &  68.89 \std{1.17} &  25.60 \std{1.91} &  64.50 \std{1.48} \\
& GPT2    &  66.90 \std{0.61} &  21.75 \std{2.16} &  62.31 \std{1.10} \\
& SCI5    &  54.07 \std{0.54} &  27.42 \std{4.77} &  51.34 \std{0.81} \\
\hline
\end{tabular}\hspace{-.1cm} %
\begin{tabular}{|ccc}
\hline
\multicolumn{3}{c}{\textbf{F1 Score}} \\
\multicolumn{1}{c}{\textbf{\seen}} &  \textbf{\unseen} &  \textbf{overall}  \\
\hline
08.02 \std{0.56} &  16.79 \std{1.30} &  10.79 \std{0.83} \\
10.43 \std{0.33} &  22.50 \std{2.83} &  13.67 \std{0.83} \\ \hline
10.18 \std{0.77} &  15.85 \std{2.01} &  12.23 \std{0.84} \\
12.20 \std{0.77} &  22.41 \std{2.96} &  15.35 \std{1.14} \\
25.56 \std{0.66} &  00.00 \std{0.00} &  14.66 \std{0.96} \\
17.50 \std{0.50} &  07.90 \std{0.63} &  13.91 \std{0.31} \\
20.80 \std{1.26} &  07.76 \std{0.49} &  17.31 \std{1.51} \\ \hline
09.61 \std{0.66} &  17.79 \std{0.70} &  12.35 \std{0.51} \\
12.71 \std{0.83} &  27.30 \std{2.13} &  16.67 \std{0.93} \\
20.89 \std{1.88} &  23.60 \std{1.39} &  22.75 \std{1.66} \\
11.36 \std{0.78} &  19.54 \std{3.04} &  13.38 \std{1.18} \\
10.86 \std{0.66} &  19.73 \std{1.98} &  13.10 \std{0.85} \\ \hline
11.38 \std{0.32} &  19.27 \std{1.27} &  14.22 \std{0.51} \\
16.11 \std{1.88} &  26.86 \std{2.79} &  19.61 \std{2.09} \\
54.19 \std{2.05} &  16.35 \std{1.59} &  39.66 \std{2.77} \\
49.29 \std{1.26} &  13.41 \std{1.28} &  35.12 \std{2.52} \\
33.36 \std{1.16} &  17.15 \std{2.70} &  27.89 \std{2.35} \\ \hline
\end{tabular}
}

\caption{Full metrics (accuracy and F1 score) of all tested models on the \prop dataset. \label{tab:full_results_prop}}
\end{table*}
\begin{table*}[htbp]
\small
\renewcommand\arraystretch{2}
\centering

{\normalsize \textbf{\coder FT} \medskip}

\resizebox{!}{1.6cm}{
\begin{tabular}{p{0cm}c|c|c|c|}
\multicolumn{2}{c}{\textbf{}} & \multicolumn{3}{c}{\textbf{Test (\seen)}} \\[-.3cm]
& \multicolumn{1}{c}{} &
\multicolumn{1}{c}{\textbf{\cadec}}        &
\multicolumn{1}{c}{\textbf{\smm}}          &
\multicolumn{1}{c}{\ \ \textbf{\prop}\ \ } \\ \cline{3-5}
\multirow{3}{*}{\rotatebox{90}{\textbf{Train}}}
& \textbf{\cadec} &39.13 &  35.80 &  31.84 \\ \cline{3-5}
& \textbf{\smm}   &  37.78 &  33.20 &  40.44 \\ \cline{3-5}
& \textbf{\prop}  &  37.83 &  35.08 &  19.88 \\ \cline{3-5}
\cline{3-5}
\end{tabular}}\ %
\resizebox{!}{1.6cm}{
\begin{tabular}{|c|c|c|}
\multicolumn{3}{c}{\textbf{Test (\unseen)}} \\[-.3cm]
\multicolumn{1}{c}{\textbf{\cadec}}        &
\multicolumn{1}{c}{\textbf{\smm}}          &
\multicolumn{1}{c}{\ \ \textbf{\prop}\ \ } \\ \cline{1-3}
45.96 &  35.79 &  08.79 \\ \cline{1-3}
31.87 &  38.91 &  09.39 \\ \cline{1-3}
35.73 &  32.56 &  28.39 \\ \cline{1-3}
\end{tabular}}\ %
\resizebox{!}{1.6cm}{
\begin{tabular}{|c|c|c|}
\multicolumn{3}{c}{\textbf{Test (overall)}} \\[-.3cm]
\multicolumn{1}{c}{\textbf{\cadec}}        &
\multicolumn{1}{c}{\textbf{\smm}}          &
\multicolumn{1}{c}{\ \ \textbf{\prop}\ \ } \\ \cline{1-3}
39.46 &  35.82 &  15.25 \\ \cline{1-3}
36.00 &  33.90 &  14.25 \\ \cline{1-3}
37.39 &  33.85 &  20.76 \\ \cline{1-3}
\end{tabular}}\\ \bigskip

{\normalsize \textbf{\coder OP+FT} \medskip}

\resizebox{!}{1.6cm}{
\begin{tabular}{p{0cm}c|c|c|c|}
\multicolumn{2}{c}{\textbf{}} & \multicolumn{3}{c}{\textbf{Test (\seen)}} \\[-.3cm]
& \multicolumn{1}{c}{} &
\multicolumn{1}{c}{\textbf{\cadec}}        &
\multicolumn{1}{c}{\textbf{\smm}}          &
\multicolumn{1}{c}{\ \ \textbf{\prop}\ \ } \\ \cline{3-5}
\multirow{3}{*}{\rotatebox{90}{\textbf{Train}}}
& \textbf{\cadec} &  42.08 &  35.12 &  33.56 \\ \cline{3-5}
& \textbf{\smm}   &  44.90 &  34.33 &  43.95 \\ \cline{3-5}
& \textbf{\prop}  &  44.25 &  38.62 &  19.64 \\ \cline{3-5}
\cline{3-5}
\end{tabular}}\ %
\resizebox{!}{1.6cm}{
\begin{tabular}{|c|c|c|}
\multicolumn{3}{c}{\textbf{Test (\unseen)}} \\[-.3cm]
\multicolumn{1}{c}{\textbf{\cadec}}        &
\multicolumn{1}{c}{\textbf{\smm}}          &
\multicolumn{1}{c}{\ \ \textbf{\prop}\ \ } \\ \cline{1-3}
48.34 &  35.63 &  11.34 \\ \cline{1-3}
36.72 &  43.24 &  12.81 \\ \cline{1-3}
37.77 &  32.21 &  32.88 \\ \cline{1-3}
\end{tabular}}\ %
\resizebox{!}{1.6cm}{
\begin{tabular}{|c|c|c|}
\multicolumn{3}{c}{\textbf{Test (overall)}} \\[-.3cm]
\multicolumn{1}{c}{\textbf{\cadec}}        &
\multicolumn{1}{c}{\textbf{\smm}}          &
\multicolumn{1}{c}{\ \ \textbf{\prop}\ \ } \\ \cline{1-3}
42.38 &  35.25 &  17.59 \\ \cline{1-3}
42.42 &  35.43 &  17.69 \\ \cline{1-3}
42.85 &  35.49 &  20.98 \\ \cline{1-3}
\end{tabular}}

\caption{Cross-dataset accuracy for \coder FT and \coder OP+FT on the three datasets.}
\label{tab:cross_coder}
\end{table*}

\begin{table*}[htbp]
\small
\renewcommand\arraystretch{2}
\centering

{\normalsize \textbf{\sapbert FT} \medskip}

\resizebox{!}{1.6cm}{
\begin{tabular}{p{0cm}c|c|c|c|}
\multicolumn{2}{c}{\textbf{}} & \multicolumn{3}{c}{\textbf{Test (\seen)}} \\[-.3cm]
& \multicolumn{1}{c}{} &
\multicolumn{1}{c}{\textbf{\cadec}}        &
\multicolumn{1}{c}{\textbf{\smm}}          &
\multicolumn{1}{c}{\ \ \textbf{\prop}\ \ } \\ \cline{3-5}
\multirow{3}{*}{\rotatebox{90}{\textbf{Train}}}
& \textbf{\cadec} &  42.69 &  35.09 &  30.80 \\ \cline{3-5}
& \textbf{\smm}   &  39.14 &  39.58 &  39.92 \\ \cline{3-5}
& \textbf{\prop}  &  32.21 &  31.07 &  26.69 \\ \cline{3-5}
\cline{3-5}
\end{tabular}}\ %
\resizebox{!}{1.6cm}{
\begin{tabular}{|c|c|c|}
\multicolumn{3}{c}{\textbf{Test (\unseen)}} \\[-.3cm]
\multicolumn{1}{c}{\textbf{\cadec}}        &
\multicolumn{1}{c}{\textbf{\smm}}          &
\multicolumn{1}{c}{\ \ \textbf{\prop}\ \ } \\ \cline{1-3}
34.47 &  35.32 &   8.73 \\ \cline{1-3}
23.79 &  32.42 &  10.02 \\ \cline{1-3}
29.77 &  34.97 &  25.81 \\ \cline{1-3}
\end{tabular}}\ %
\resizebox{!}{1.6cm}{
\begin{tabular}{|c|c|c|}
\multicolumn{3}{c}{\textbf{Test (overall)}} \\[-.3cm]
\multicolumn{1}{c}{\textbf{\cadec}}        &
\multicolumn{1}{c}{\textbf{\smm}}          &
\multicolumn{1}{c}{\ \ \textbf{\prop}\ \ } \\ \cline{1-3}
42.30 &  35.12 &  14.94 \\ \cline{1-3}
34.53 &  38.69 &  14.71 \\ \cline{1-3}
31.71 &  32.98 &  26.60 \\ \cline{1-3}
\end{tabular}}\\ \bigskip

{\normalsize \textbf{\sapbert OP+FT} \medskip}

\resizebox{!}{1.6cm}{
\begin{tabular}{p{0cm}c|c|c|c|}
\multicolumn{2}{c}{\textbf{}} & \multicolumn{3}{c}{\textbf{Test (\seen)}} \\[-.3cm]
& \multicolumn{1}{c}{} &
\multicolumn{1}{c}{\textbf{\cadec}}        &
\multicolumn{1}{c}{\textbf{\smm}}          &
\multicolumn{1}{c}{\ \ \textbf{\prop}\ \ } \\ \cline{3-5}
\multirow{3}{*}{\rotatebox{90}{\textbf{Train}}}
& \textbf{\cadec} & 63.96 &  45.93 &  49.29 \\ \cline{3-5}
& \textbf{\smm}   & 70.71 &  46.28 &  62.73 \\ \cline{3-5}
& \textbf{\prop}  & 68.69 &  49.77 &  41.33 \\ \cline{3-5}
\cline{3-5}
\end{tabular}}\ %
\resizebox{!}{1.6cm}{
\begin{tabular}{|c|c|c|}
\multicolumn{3}{c}{\textbf{Test (\unseen)}} \\[-.3cm]
\multicolumn{1}{c}{\textbf{\cadec}}        &
\multicolumn{1}{c}{\textbf{\smm}}          &
\multicolumn{1}{c}{\ \ \textbf{\prop}\ \ } \\ \cline{1-3}
47.76 &  34.72 &  14.63 \\ \cline{1-3}
61.31 &  48.43 &  17.84 \\ \cline{1-3}
48.09 &  28.82 &  39.81 \\ \cline{1-3}
\end{tabular}}\ %
\resizebox{!}{1.6cm}{
\begin{tabular}{|c|c|c|}
\multicolumn{3}{c}{\textbf{Test (overall)}} \\[-.3cm]
\multicolumn{1}{c}{\textbf{\cadec}}        &
\multicolumn{1}{c}{\textbf{\smm}}          &
\multicolumn{1}{c}{\ \ \textbf{\prop}\ \ } \\ \cline{1-3}
63.19 &  42.65 &  24.39 \\ \cline{1-3}
67.89 &  46.54 &  24.90 \\ \cline{1-3}
64.24 &  39.45 &  41.17 \\ \cline{1-3}
\end{tabular}}

\caption{Cross-dataset accuracy for \sapbert FT and \sapbert OP+FT on the three datasets.}
\label{tab:cross_sapbert}
\end{table*}

\begin{table*}[htbp]
\small
\renewcommand\arraystretch{2}
\centering

{\normalsize \textbf{\pubmedbert FT} \medskip}

\resizebox{!}{1.6cm}{
\begin{tabular}{p{0cm}c|c|c|c|}
\multicolumn{2}{c}{\textbf{}} & \multicolumn{3}{c}{\textbf{Test (\seen)}} \\[-.3cm]
& \multicolumn{1}{c}{} &
\multicolumn{1}{c}{\textbf{\cadec}}        &
\multicolumn{1}{c}{\textbf{\smm}}          &
\multicolumn{1}{c}{\ \ \textbf{\prop}\ \ } \\ \cline{3-5}
\multirow{3}{*}{\rotatebox{90}{\textbf{Train}}}
& \textbf{\cadec} & 82.47 &  54.00 &  51.74 \\ \cline{3-5}
& \textbf{\smm}   & 38.82 &  50.26 &  31.43 \\ \cline{3-5}
& \textbf{\prop}  & 47.76 &  39.83 &  60.20 \\ \cline{3-5}
\cline{3-5}
\end{tabular}}\ %
\resizebox{!}{1.6cm}{
\begin{tabular}{|c|c|c|}
\multicolumn{3}{c}{\textbf{Test (\unseen)}} \\[-.3cm]
\multicolumn{1}{c}{\textbf{\cadec}}        &
\multicolumn{1}{c}{\textbf{\smm}}          &
\multicolumn{1}{c}{\ \ \textbf{\prop}\ \ } \\ \cline{1-3}
00.00 &    00.00 &   00.00 \\ \cline{1-3}
00.00 &    00.00 &   00.00 \\ \cline{1-3}
00.00 &    00.00 &   00.00 \\ \cline{1-3}
\end{tabular}}\ %
\resizebox{!}{1.6cm}{
\begin{tabular}{|c|c|c|}
\multicolumn{3}{c}{\textbf{Test (overall)}} \\[-.3cm]
\multicolumn{1}{c}{\textbf{\cadec}}        &
\multicolumn{1}{c}{\textbf{\smm}}          &
\multicolumn{1}{c}{\ \ \textbf{\prop}\ \ } \\ \cline{1-3}
78.49 &  38.00 &  14.55 \\ \cline{1-3}
27.16 &  44.12 &  04.93 \\ \cline{1-3}
37.48 &  20.22 &  54.09 \\ \cline{1-3}
\end{tabular}}\\ \bigskip

{\normalsize \textbf{\pubmedbert OP+FT} \medskip}

\resizebox{!}{1.6cm}{
\begin{tabular}{p{0cm}c|c|c|c|}
\multicolumn{2}{c}{\textbf{}} & \multicolumn{3}{c}{\textbf{Test (\seen)}} \\[-.3cm]
& \multicolumn{1}{c}{} &
\multicolumn{1}{c}{\textbf{\cadec}}        &
\multicolumn{1}{c}{\textbf{\smm}}          &
\multicolumn{1}{c}{\ \ \textbf{\prop}\ \ } \\ \cline{3-5}
\multirow{3}{*}{\rotatebox{90}{\textbf{Train}}}
& \textbf{\cadec} & 87.18 &  61.05 &  64.95 \\ \cline{3-5}
& \textbf{\smm}   & 79.84 &  74.10 &  79.24 \\ \cline{3-5}
& \textbf{\prop}  & 75.33 &  63.33 &  68.89 \\ \cline{3-5}
\cline{3-5}
\end{tabular}}\ %
\resizebox{!}{1.6cm}{
\begin{tabular}{|c|c|c|}
\multicolumn{3}{c}{\textbf{Test (\unseen)}} \\[-.3cm]
\multicolumn{1}{c}{\textbf{\cadec}}        &
\multicolumn{1}{c}{\textbf{\smm}}          &
\multicolumn{1}{c}{\ \ \textbf{\prop}\ \ } \\ \cline{1-3}
23.88 &  18.76 &  11.64 \\ \cline{1-3}
53.14 &  37.46 &  17.35 \\ \cline{1-3}
39.84 &  23.86 &  25.60 \\ \cline{1-3}
\end{tabular}}\ %
\resizebox{!}{1.6cm}{
\begin{tabular}{|c|c|c|}
\multicolumn{3}{c}{\textbf{Test (overall)}} \\[-.3cm]
\multicolumn{1}{c}{\textbf{\cadec}}        &
\multicolumn{1}{c}{\textbf{\smm}}          &
\multicolumn{1}{c}{\ \ \textbf{\prop}\ \ } \\ \cline{1-3}
84.12 &  48.54 &  26.64 \\ \cline{1-3}
71.82 &  69.64 &  27.07 \\ \cline{1-3}
67.72 &  43.88 &  64.50 \\ \cline{1-3}
\end{tabular}}

\caption{Cross-dataset accuracy for \pubmedbert FT and \pubmedbert OP+FT on the three datasets.}
\label{tab:cross_pmb}
\end{table*}

\begin{table*}[htbp]
\small
\renewcommand\arraystretch{2}
\centering

{\normalsize \textbf{\gpt FT} \medskip}

\resizebox{!}{1.6cm}{
\begin{tabular}{p{0cm}c|c|c|c|}
\multicolumn{2}{c}{\textbf{}} & \multicolumn{3}{c}{\textbf{Test (\seen)}} \\[-.3cm]
& \multicolumn{1}{c}{} &
\multicolumn{1}{c}{\textbf{\cadec}}        &
\multicolumn{1}{c}{\textbf{\smm}}          &
\multicolumn{1}{c}{\ \ \textbf{\prop}\ \ } \\ \cline{3-5}
\multirow{3}{*}{\rotatebox{90}{\textbf{Train}}}
& \textbf{\cadec} & 84.70 &  60.86 &  60.52 \\ \cline{3-5}
& \textbf{\smm}   & 65.73 &  72.31 &  69.88 \\ \cline{3-5}
& \textbf{\prop}  & 56.68 &  52.43 &  65.21 \\ \cline{3-5}
\cline{3-5}
\end{tabular}}\ %
\resizebox{!}{1.6cm}{
\begin{tabular}{|c|c|c|}
\multicolumn{3}{c}{\textbf{Test (\unseen)}} \\[-.3cm]
\multicolumn{1}{c}{\textbf{\cadec}}        &
\multicolumn{1}{c}{\textbf{\smm}}          &
\multicolumn{1}{c}{\ \ \textbf{\prop}\ \ } \\ \cline{1-3}
07.40 &  03.78 &  01.51 \\ \cline{1-3}
06.54 &  10.02 &  01.84 \\ \cline{1-3}
07.41 &  03.45 &  05.75 \\ \cline{1-3}
\end{tabular}}\ %
\resizebox{!}{1.6cm}{
\begin{tabular}{|c|c|c|}
\multicolumn{3}{c}{\textbf{Test (overall)}} \\[-.3cm]
\multicolumn{1}{c}{\textbf{\cadec}}        &
\multicolumn{1}{c}{\textbf{\smm}}          &
\multicolumn{1}{c}{\ \ \textbf{\prop}\ \ } \\ \cline{1-3}
80.97 &  43.99 &  18.10 \\ \cline{1-3}
47.94 &  64.68 &  12.52 \\ \cline{1-3}
46.08 &  28.31 &  59.18 \\ \cline{1-3}
\end{tabular}}\\ \bigskip

{\normalsize \textbf{\gpt OP+FT} \medskip}

\resizebox{!}{1.6cm}{
\begin{tabular}{p{0cm}c|c|c|c|}
\multicolumn{2}{c}{\textbf{}} & \multicolumn{3}{c}{\textbf{Test (\seen)}} \\[-.3cm]
& \multicolumn{1}{c}{} &
\multicolumn{1}{c}{\textbf{\cadec}}        &
\multicolumn{1}{c}{\textbf{\smm}}          &
\multicolumn{1}{c}{\ \ \textbf{\prop}\ \ } \\ \cline{3-5}
\multirow{3}{*}{\rotatebox{90}{\textbf{Train}}}
& \textbf{\cadec} & 86.08 &  62.50 &  62.56 \\ \cline{3-5}
& \textbf{\smm}   & 78.23 &  73.71 &  77.98 \\ \cline{3-5}
& \textbf{\prop}  & 72.75 &  58.96 &  66.90 \\ \cline{3-5}
\cline{3-5}
\end{tabular}}\ %
\resizebox{!}{1.6cm}{
\begin{tabular}{|c|c|c|}
\multicolumn{3}{c}{\textbf{Test (\unseen)}} \\[-.3cm]
\multicolumn{1}{c}{\textbf{\cadec}}        &
\multicolumn{1}{c}{\textbf{\smm}}          &
\multicolumn{1}{c}{\ \ \textbf{\prop}\ \ } \\ \cline{1-3}
23.92 &  34.21 &   08.90 \\ \cline{1-3}
26.72 &  31.25 &  11.02 \\ \cline{1-3}
27.65 &  18.72 &  21.75 \\ \cline{1-3}
\end{tabular}}\ %
\resizebox{!}{1.6cm}{
\begin{tabular}{|c|c|c|}
\multicolumn{3}{c}{\textbf{Test (overall)}} \\[-.3cm]
\multicolumn{1}{c}{\textbf{\cadec}}        &
\multicolumn{1}{c}{\textbf{\smm}}          &
\multicolumn{1}{c}{\ \ \textbf{\prop}\ \ } \\ \cline{1-3}
83.08 &  54.06 &  24.00 \\ \cline{1-3}
62.74 &  68.52 &  21.54 \\ \cline{1-3}
63.09 &  39.15 &  62.31 \\ \cline{1-3}
\end{tabular}}

\caption{Cross-dataset accuracy for \gpt FT and \gpt OP+FT on the three datasets.}
\label{tab:cross_gpt}
\end{table*}

\begin{table*}[htbp]
\small
\renewcommand\arraystretch{2}
\centering

{\normalsize \textbf{\scifive FT} \medskip}

\resizebox{!}{1.6cm}{
\begin{tabular}{p{0cm}c|c|c|c|}
\multicolumn{2}{c}{\textbf{}} & \multicolumn{3}{c}{\textbf{Test (\seen)}} \\[-.3cm]
& \multicolumn{1}{c}{} &
\multicolumn{1}{c}{\textbf{\cadec}}        &
\multicolumn{1}{c}{\textbf{\smm}}          &
\multicolumn{1}{c}{\ \ \textbf{\prop}\ \ } \\ \cline{3-5}
\multirow{3}{*}{\rotatebox{90}{\textbf{Train}}}
& \textbf{\cadec} & 72.80 &  44.44 &  46.20 \\ \cline{3-5}
& \textbf{\smm}   & 35.28 &  34.73 &  44.05 \\ \cline{3-5}
& \textbf{\prop}  & 29.23 &  28.44 &  45.70 \\ \cline{3-5}
\cline{3-5}
\end{tabular}}\ %
\resizebox{!}{1.6cm}{
\begin{tabular}{|c|c|c|}
\multicolumn{3}{c}{\textbf{Test (\unseen)}} \\[-.3cm]
\multicolumn{1}{c}{\textbf{\cadec}}        &
\multicolumn{1}{c}{\textbf{\smm}}          &
\multicolumn{1}{c}{\ \ \textbf{\prop}\ \ } \\ \cline{1-3}
15.94 &  06.86 &  03.07 \\ \cline{1-3}
10.42 &  12.63 &  03.30 \\ \cline{1-3}
15.44 &  11.14 &  12.71 \\ \cline{1-3}
\end{tabular}}\ %
\resizebox{!}{1.6cm}{
\begin{tabular}{|c|c|c|}
\multicolumn{3}{c}{\textbf{Test (overall)}} \\[-.3cm]
\multicolumn{1}{c}{\textbf{\cadec}}        &
\multicolumn{1}{c}{\textbf{\smm}}          &
\multicolumn{1}{c}{\ \ \textbf{\prop}\ \ } \\ \cline{1-3}
70.05 &  33.33 &  15.19 \\ \cline{1-3}
27.81 &  32.06 &  09.71 \\ \cline{1-3}
26.24 &  19.93 &  42.34 \\ \cline{1-3}
\end{tabular}}\\ \bigskip

{\normalsize \textbf{\scifive OP+FT} \medskip}

\resizebox{!}{1.6cm}{
\begin{tabular}{p{0cm}c|c|c|c|}
\multicolumn{2}{c}{\textbf{}} & \multicolumn{3}{c}{\textbf{Test (\seen)}} \\[-.3cm]
& \multicolumn{1}{c}{} &
\multicolumn{1}{c}{\textbf{\cadec}}        &
\multicolumn{1}{c}{\textbf{\smm}}          &
\multicolumn{1}{c}{\ \ \textbf{\prop}\ \ } \\ \cline{3-5}
\multirow{3}{*}{\rotatebox{90}{\textbf{Train}}}
& \textbf{\cadec} & 80.96 &  56.07 &  60.35 \\ \cline{3-5}
& \textbf{\smm}   & 71.96 &  60.76 &  72.82 \\ \cline{3-5}
& \textbf{\prop}  & 63.63 &  47.74 &  54.07 \\ \cline{3-5}
\cline{3-5}
\end{tabular}}\ %
\resizebox{!}{1.6cm}{
\begin{tabular}{|c|c|c|}
\multicolumn{3}{c}{\textbf{Test (\unseen)}} \\[-.3cm]
\multicolumn{1}{c}{\textbf{\cadec}}        &
\multicolumn{1}{c}{\textbf{\smm}}          &
\multicolumn{1}{c}{\ \ \textbf{\prop}\ \ } \\ \cline{1-3}
37.21 &  36.86 &  13.95 \\ \cline{1-3}
52.17 &  37.87 &  17.19 \\ \cline{1-3}
40.12 &  26.95 &  27.42 \\ \cline{1-3}
\end{tabular}}\ %
\resizebox{!}{1.6cm}{
\begin{tabular}{|c|c|c|}
\multicolumn{3}{c}{\textbf{Test (overall)}} \\[-.3cm]
\multicolumn{1}{c}{\textbf{\cadec}}        &
\multicolumn{1}{c}{\textbf{\smm}}          &
\multicolumn{1}{c}{\ \ \textbf{\prop}\ \ } \\ \cline{1-3}
78.85 &  50.34 &  27.02 \\ \cline{1-3}
66.02 &  57.98 &  25.92 \\ \cline{1-3}
58.52 &  37.45 &  51.34 \\ \cline{1-3}
\end{tabular}}

\caption{Cross-dataset accuracy for \scifive FT and \scifive OP+FT on the three datasets.}
\label{tab:cross_scifive}
\end{table*}

\end{document}